\documentclass[journal ]{new-aiaa}
\usepackage[utf8]{inputenc}
\usepackage{textcomp}

\usepackage{graphicx}
\usepackage{amsmath}
\usepackage[version=4]{mhchem}
\usepackage{siunitx}
\usepackage{cases}
\usepackage{longtable,tabularx}
\usepackage{hyperref}
\usepackage{epstopdf}
\hypersetup{
	colorlinks=true,
	linkcolor=blue,
	filecolor=magenta,      
	urlcolor=cyan,
	citecolor=black,
}
\title{An Image Based Visual Servo Method for Probe-and-Drogue Autonomous Aerial Refueling}

\author{Quan Quan\footnote{Professor, School of Automation Science and Electrical Engineering, qq\_buaa@buaa.edu.cn}, Runxiao Liu, \footnote{PhD, Candidate, School of Automation Science and Electrical Engineering, runxiao\_liu@buaa.edu.cn.} Hao Liu, \footnote{Master, Candidate, School of Automation Science and Electrical Engineering, zy1903267@buaa.edu.cn.}  Zeqing Ma, \footnote{Master, School of Automation Science and Electrical Engineering, 610192064@qq.com.}}
\affil{Beihang University, 100191 Beijing, People's Republic of China}
\author{ Jinrui Ren, \footnote{Associate Professor,School of Automation, renjinrui@xupt.edu.cn} }
\affil{Xi’an University of Posts {\&} Telecommunications, 710121 Xi'an, Shaanxi, People's Republic of China}

\begin{document}

\maketitle

\section{Introduction}
Autonomous aerial refueling (AAR) is an important capability for the future successful deployment and operation of unmanned aerial vehicles (UAVs) \cite{nalepka2005automated}. As a widely-used aerial refueling method, the probe-drogue refueling (PDR) system is considered to be more flexible and compact than other aerial refueling systems. However, a drawback of PDR is that the drogue is passive and susceptible to aerodynamic disturbances \cite{THOMAS201414,DAI2016448,7738351}. What is more, a key issue is represented by the need of a highly accurate measurement of the relative ‘tanker-receiver’ position and orientation in the final phase of docking and during the refueling.  Therefore, it is difficult to design a reliable and robust controller for the receiver aircraft.  Various sensing technologies have been employed, including inertial measurements, differential GPS (DGPS), and electro-optical systems \cite{THOMAS201414,martinez2013vision}. Although these sensors are suitable for autonomous docking, there might be limitations associated with their use. For example, receiver's GPS signals might not always be available when they are shadowed or distorted by the tanker. Therefore, the use of machine vision technologies have been proposed in addition – or as an alternative – to these technologies. The combination of GPS measurements with position estimates from vision systems has been explored in Ref. \cite{campa2004autonomous}, and some vision-based navigation systems can be found in Refs. \cite{martinez2013vision,valasek2005vision,wang2015real}.

Visual servo schemes mainly differ in the way the features are designed. The existing visual servo control system schemes are classified as position-based visual servos (PBVS) and image-based visual servos (IBVS). In PBVS, the features consist of a set of 3D parameters which must be estimated from image measurements, and it is typical to define the features in terms of the pose of the camera with respect to some reference frame \cite{chaumette2006visual}. The most commonly-used method uses a set of infrared light emitting diodes (LEDs) \cite{wang2015real} on the parachute part of a drogue as marks, and then uses the Gaussian least-square-differential-correction (GLSDC) algorithm \cite{valasek2005vision} or a deep leaning method \cite{liu2018deep} to obtain relative pose. In IBVS, the features consist of a set of 2D features that are immediately available in the image data \cite{chaumette2006visual}. 
During the past decades, the visual servo control method has been applied to AAR systems by many researchers. These studies mainly focus on pose estimation methods and controller design, in order to allow the receiver to determine the relative position of the refueling drogue and control strategies, to enable a robust and safe approach and docking. Recently, control methods like iterative learning control \cite{dai2018terminal,ren2019reliable} and fault-tolerant adaptive model inversion control \cite{valasek2017fault} have received much attention.

On the whole, there still exist some challenges in visual servo control for AAR. Firstly, the PDR system is complicated. As one of the main disturbances in the docking stage, the bow wave effect is a state-dependent nonlinear disturbance which affects the docking process \cite{DAI2016448,7738351}. Moreover, it should be noticed that an accurate 2D image observation does not implies an accurate 3D pose estimation due to the camera calibration error, installation error and 3D object modeling error. Most recent works \cite{valasek2017fault,dai2018terminal,ren2019reliable} are based on PBVS, for which an accurate 3D pose estimation is crucial, since it appears both in the error to be regulated to 0 and in interaction matrix. Coarse estimation will thus cause perturbations on the trajectory realized but will have also an effect on the accuracy of the pose reached. 

A reliable docking control method based on IBVS for PDR is proposed in this paper to cope with the challenges mentioned above. Compared with PBVS, in IBVS, 2D image error is used to make the aircraft reach its desired pose directly. What is more, the linear quadratic regulator (LQR)
method is used here to reject complex disturbances and uncertainties which comprises the bow wave effect during the docking stage. 
The main contributions of this paper are as follows.

\begin{itemize}
	\item An IBVS model with a forward-looking monocular camera mounted on the receiver is established first. The model combines the underactuated characteristics of the receiver's motion with the Jacobian matrix related to the derivative of the image space measurements to the camera's linear and angular velocities.
	\item Based on the established model, the inner and outer loop controllers are designed. The outer loop makes the 2D image error converge to zero, and the inner loop is using the LQR method to get the optimal inputs while rejecting complex disturbances. The proposed control method is robust against to the pose estimation error.
	\item  Simulations are carried out to validate the effectiveness of the proposed methods subject to aerodynamic disturbances, including the bow wave effect. Besides, a comparison is made between IBVS and PBVS subject to pose estimation error.
\end{itemize}

This paper is organized as follows. The model description and problem statement are introduced in Sec. \uppercase\expandafter{\romannumeral2} to transform the practical docking control problem into an IBVS control problem. In Sec. \uppercase\expandafter{\romannumeral3}, the main algorithm used in this paper is presented. Then, in Sec. \uppercase\expandafter{\romannumeral4}, the details and results of the simulation are presented. Finally, in Sec. \uppercase\expandafter{\romannumeral5}, the conclusions are drawn.

\section{System Description and Problem Formulation}
In order to describe the docking control problem, basic coordinate frames and aircraft models are introduced, and then, the control problem is formulated.
\subsection{PDR system model in docking stage}
The docking control task for the receiver aircraft is to approach the tanker and then dock the fuel probe tip with the drogue receptacle. When establishing the PDR system model in the docking stage, five coordinate frames needed are the ground coordinate frame ($ o_{\rm{g}}x_{\rm{g}}y_{\rm{g}}z_{\rm{g}} $), the receiver coordinate frame ($ o_{\rm{r}}x_{\rm{r}}y_{\rm{r}}z_{\rm{r}} $), the tanker coordinate frame ($ o_{\rm{t}}x_{\rm{t}}y_{\rm{t}}z_{\rm{t}} $), the camera coordinate frame ($ o_{\rm{c}}x_{\rm{c}}y_{\rm{c}}z_{\rm{c}} $) and a reference coordinate frame ($ o_{\rm{re}}x_{\rm{re}}y_{\rm{re}}z_{\rm{re}} $), which are shown in Fig. 1. Specially, the origin of the reference frame coincides with the tanker coordinate frame, and its coordinate axis direction  is consistent with the camera coordinate frame.  The relative position and attitude between the drogue and the receiver are mainly concerned in the docking phase,
where the tanker frame (moving inertial frame) is more convenient for modeling and analysis than the ground frame
(fixed inertial frame). In order to further simplify the coordinate transformation among different frames, the direction
of the axis $ o_{\rm{g}}x_{\rm{g}} $ of the ground frame is also selected the same as the horizontal moving direction of the tanker. Besides, $ \textbf{R}_{\rm{r/t}} $ denotes the  \textit{rotation matrix} to describe the angular relationship
of the receiver frame ``r'' relative to the tanker frame ``t''. The symbolic operation rules here are defined as
\begin{equation*} \left\{\begin{matrix}
	\mathbf{R}_{{i/j}}\textbf{x}_{{k}}^{{j}}=\textbf{x}_{{k}}^{{i}} \\
	\textbf{x}_{{i}}^{{j}}-\textbf{x}_{{k}}^{{j}}=\textbf{x}_{{k}}^{{i}}
\end{matrix} \right.
\end{equation*}
where $ {i},{j} $ denotes which coordinate system it is in and  $ {k} $ represents which object it is belong to. For instance, $ {V}^{\rm{g}}_{\rm{t}} $ is the forward flight velocity of the \textit{t}anker in the \textit{g}round frame and $ \textbf{p}_{\rm{c}}^{\rm{r}} $ is the difference vector from the \textit{r}eceiver to the \textit{c}amera, namely $ \textbf{p}_{\rm{r}}^{\rm{c}}=\textbf{p}_{\rm{c}}^{\rm{t}}- \textbf{p}^{\rm{t}}_{\rm{r}}$ in Fig. 1.
  The formal definition of these coordinate frames can be found in Ref. \cite{7738351}.
\begin{figure}[hbt!]
	\centering
	\includegraphics[width=.8\textwidth]{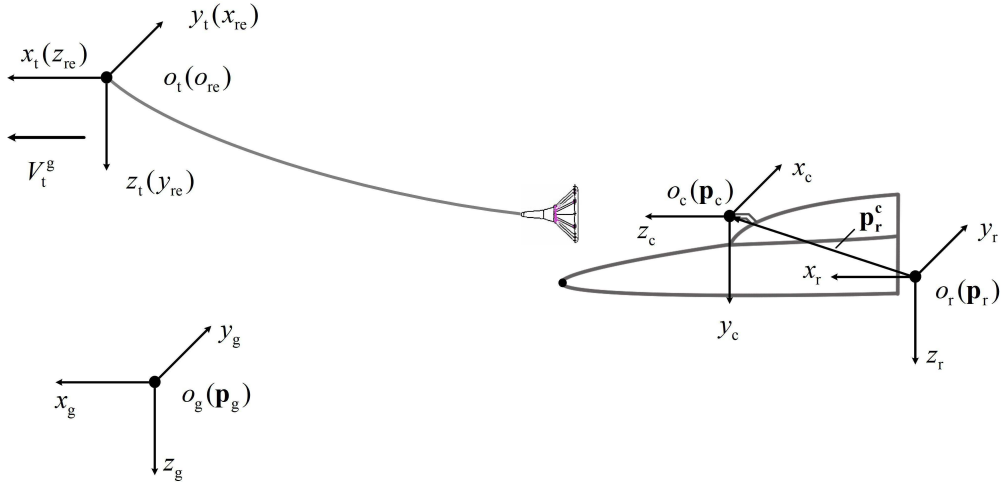}
	\caption{Coordinate frames used in the docking process}
\end{figure}

During the docking stage, after trimming and linearization, the equations of motion of the receiver \cite{stevens2015aircraft} under the tanker coordinate frame can be simply represented as

\begin{equation}
	\left\{ \begin{aligned}
		\boldsymbol{\dot{\tilde{\rm{x}} }} _{\rm{rlon}} &=\rm{\mathbf{A} }_{rlon} \tilde{\boldsymbol{\rm{x}} } _{\rm{rlon}}+\rm{\mathbf{B} }_{rlon}\tilde{\boldsymbol{\rm{u}} } _{\rm{rlon}}  \\  
		\dot{\tilde{\boldsymbol{\rm{x}} } } _{\rm{rlat}} &=\rm{\mathbf{A} }_{rlat} \tilde{\boldsymbol{\rm{x}} } _{\rm{rlat}}+\rm{\mathbf{B} }_{rlat}\tilde{\boldsymbol{\rm{u}} } _{\rm{rlat}} 
	\end{aligned}\right. 
\end{equation}
where $ \tilde{\mathbf{u}}\rm{_{rlon}}=$ [ $  \tilde{\delta}_{\rm{a}} $   $  \tilde{\delta}_{\rm{r}}  $ ]
$ ^{\text{T}} $
and $ \tilde{\mathbf{u}}\rm{_{rlat}}=$[ $ \tilde{\delta}_{\rm{e}} $   $ \tilde{\delta}_{\rm{t}}  $ ]
$ ^{\text{T}} $ denote the control input, which
comprises the aileron,rudder, elevator and throttle,  respectively. The vector $ \tilde{\boldsymbol{\rm{x} } } _{\rm{rlon}} =$ [ $ \tilde{x}_{\rm{r}} $  $ \tilde{h}_{\rm{r}} $   $ \tilde{\theta} $ $ \tilde{{V}}^{\rm{g}}_{\rm{r}} $ $ \tilde{\alpha} $   $  \tilde{q} $ ]$ ^{\rm{T}} $ and $ \tilde{\boldsymbol{\rm{x}} }_{\rm{rlat}}=$ [ $ \tilde{y}_{\rm{r}} $ $ \tilde{\psi} $  $ \tilde{\phi} $   $ \tilde{\beta} $  $ \tilde{p} $ $ \tilde{r} $ ] $ ^{\rm{T}} $  denote the state of the receiver, which comprises position, Euler angles, aerodynamic angles and angular rate, respectively. Here, the subscript ``rlon'' and ``rlat'' refer to the longitudinal and lateral channel of the receiver, respectively. Note that, all the system matrices $ \textbf{A}\rm{_{rlon}} $, $\textbf{A}\rm{_{rlat}} $, $ \textbf{B}\rm{_{rlon}} $, $ \textbf{B}\rm{_{rlat}} $ are known time-invariant matrices with appropriate dimensions and the symbol "$ \sim $"  means that the model should do incremental control on the  trimming state.

\subsection{Problem Formulation}
\begin{figure}[hbt!]
	\centering
	\includegraphics[width=.5\textwidth]{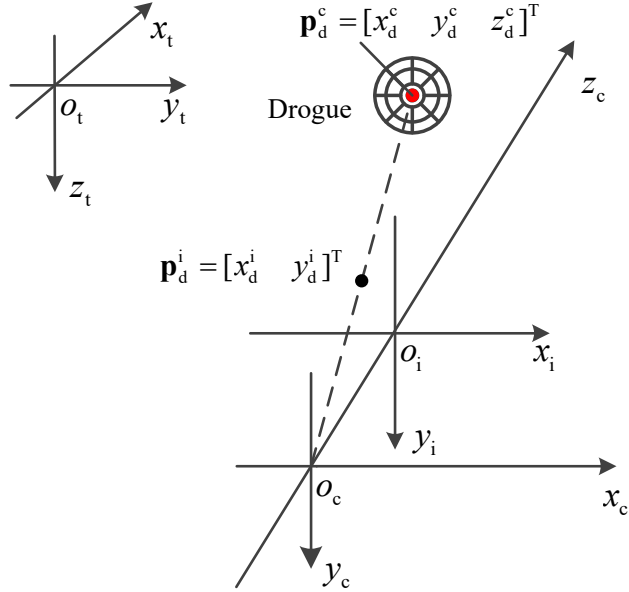}
	\caption{Relationship among coordinate frames. $ o_{\rm{t}}x_{\rm{t}}y_{\rm{t}}z_{\rm{t}}$ is the tanker coordinate frame $ {\rm{t}} $; $ o_{\rm{c}}x_{\rm{c}}y_{\rm{c}}z_{\rm{c}}$ is the camera coordinate frame $ {\rm{c}} $, and the conversion from  $ {\rm{t}} $ to  $ {\rm{c}} $ is  $ \mathbf{R} \rm{_{c/t}} $; $ o_{\rm{i}}x_{\rm{i}}y_{\rm{i}}$ is the image coordinate  frame $ {\rm{i}} $.}
\end{figure}
Define the coordinate of the drogue's center  in the camera coordinate frame as $ \mathbf{p}^{\rm{c}}_{\rm{d}}= $ [ $ x^{\rm{c}}_{\rm{d}} $  $ y^{\rm{c}}_{\rm{d}} $ $ z^{\rm{c}}_{\rm{d}} $ ]$ ^{\rm{T}} $, which is projected in the image as a 2D point with coordinates $ \mathbf{p}^{\rm{i}}_{\rm{d}} =$ [ $ x^{\rm{i}}_{\rm{d}} $ $ y^{\rm{i}}_{\rm{d}} $ ] $ ^{\rm{T}}$ as shown in Fig. 2. 
The image tracking error $ \textbf{e} $ in the image is defined as 
\begin{equation}
	\mathbf{e}=\begin{bmatrix}
		e_{x} \\
		e_{y} 
	\end{bmatrix}=\begin{bmatrix}
		x^{\rm{i}}_{\rm{d}}-x_{\rm{o}}^{\rm{i}} \\
		y^{\rm{i}}_{\rm{d}}-y_{\rm{o}}^{\rm{i}}
	\end{bmatrix}
		\end{equation}
where $ \left (x_{\rm{o}}^{\rm{i}}, y_{\rm{o}}^{\rm{i}}\right )$ is a fixed point in the image called the convergence point of image, which is the origin in the image frame here. Before proceeding further, three assumptions are made in the following.

\textbf{Assumption 1.}
	The visual tracking module of the receiver is able to accurately capture the drogue's central coordinate $  \textbf{p}^{\rm{i}}_{\rm{d}} $ from images in real time.

\textbf{Assumption 2.}
    The tanker is in a steady flight during the entire docking process, while the receiver keeps a certain distance to the tanker with the trim state $ \textbf{v}_{\rm{r}}^{\rm{\ast}}= \textbf{v}_{\rm{t}}^{\rm{\ast}}$.

\textbf{Assumption 3.}    
 The camera installation position coincides with the head of the probe as shown in Fig. 1. The transfer matrix from the tanker coordinate frame to camera coordinate frame is 
 \begin{equation}
 	\mathbf{R} \rm{_{c/t}}=\begin{bmatrix}
 		0 &1  &0 \\
 		0&0 & 1\\
 		1 & 0 &0
 	\end{bmatrix}.
 \end{equation}

Based on \textit{Assumptions 1-3}, the \textit{Visual Servo Control Problem} is stated in the following.

 The problem is to design properly input $ \tilde{\mathbf{u}}\rm{_{r}}=$ [ $  \tilde{\delta}_{\rm{a}} $   $  \tilde{\delta}_{\rm{r}}  $ $ \tilde{\delta}_{\rm{e}} $  $ \tilde{\delta}_{\rm{t}}$ ]
 $ ^{\text{T}} $
for system (1) to make the image tracking error converge to zero $ \left ( \mathbf{e}(t) \to 0 \right )  $ and depth error converging to zero $ \left ( z^{\rm{c}}_{\rm{d}}(t) \to 0 \right )$ as $ t \to  \infty$.
\section{IBVS controller design}
No matter which kind of method is used (IBVS, PBVS, or homography-based visual servo control), it is supposed
to calculate desired linear velocity and angular velocity (six control variables) through the motion of image features. However, the receiver is an underactuated system with six degrees of freedom controlled by four control inputs. Therefore, the six degrees of freedom of the receiver are coupled, which is the obstacle while using IBVS. In this section, an IBVS model with a forward-looking monocular camera mounted on the receiver is established first with a longitudinal channel model and a lateral channel model derived. Besides, based on the two obtained models, the controllers are designed separately.

\subsection{Visual servo model based on the Jacobian matrix}

According to the basic equation of IBVS \cite{chaumette2006visual}, the relationship between $ \dot{\mathbf{e}}  $ and $ \textbf{v}^{\rm{c}}_{\rm{d}} $, $ \boldsymbol{\omega}^{\rm{c}}_{\rm{d}} $ is
 \begin{equation}
 	\dot{\mathbf{e}}=\underbrace{\begin{bmatrix}
 		-\frac{1}{z_{\rm{d}}^{\rm{c}} } & 0 &\frac{x^{\rm{i}} }{z_{\rm{d}}^{\rm{c}} }   & x^{\rm{i}}y^{\rm{i}}   & -(1+{x^{\rm{i}}}^{2} ) & y^{\rm{i}} \\
 		0& -\frac{1}{z_{\rm{d}}^{\rm{c}}}  & \frac{y^{\rm{i}} }{z_{\rm{d}}^{\rm{c}}}  &1+{y^{\rm{i}}}^{2}   &-x^{\rm{i}}y^{\rm{i}}    &-x^{\rm{i}}
 	\end{bmatrix}}_{\mathbf{L}} \left [\begin{matrix} v^{\rm{c}}_{x} \\  v^{\rm{c}}_{y} \\ v^{\rm{c}}_{z} \\ \omega^{\rm{c}}_{x} \\ \omega^{\rm{c}}_{y} \\ \omega^{\rm{c}}_{z}\end{matrix}\right ]
 \end{equation}
where
$ \mathbf{v}_{\rm{d}}^{\rm{c}}=$ [ $ v^{\rm{c}}_{{\rm{d}},x} $ $ v^{\rm{c}}_{{\rm{d}},y} $ $ v^{\rm{c}}_{{\rm{d}},z} $ ]$ ^{\rm{T}} $ $=\mathbf{R}_{\rm{re/t}}(\textbf{v}_{\rm{c}}^{\rm{t}}-\textbf{v}_{\rm{d}}^{\rm{t}})=\textbf{v}_{\rm{c}}^{\rm{re}}-\textbf{v}_{\rm{d}}^{\rm{re}}$ is the relative  velocity of the drogue and $ \boldsymbol{\omega}_{\rm{d}}^{\rm{c}}=$ [ $ \omega^{\rm{c}}_{{\rm{d}},x} $ $ \omega^{\rm{c}}_{{\rm{d}},y} $ $ \omega^{\rm{c}}_{{\rm{d}},z} $]$ ^{\rm{T}} $ is the angular velocity of the drogue in the camera coordinate frame . Specially, $ \mathbf{R}_{\rm{re/t}}=  \mathbf{R}_{\rm{c/t}}$ is the transfer matrix from the tanker coordinate frame to the reference coordinate frame, and $ \textbf{v}_{\rm{c}}^{\rm{re}} =$ [ $ v^{\rm{re}}_{{\rm{c}},x} $ $ v^{\rm{re}}_{{\rm{c}},y} $ $ v^{\rm{re}}_{{\rm{c}},z} $ ]$ ^{\rm{T}} $  and $ \textbf{v}_{\rm{d}}^{\rm{re}} $ are the velocity of the camera and the drogue under the reference coordinate frame. Furthermore, $\textbf{L} \in {\mathbb{R}^{2 \times 6}}$ is called the Jacobian matrix, $ z^{\rm{c}}_{\rm{d}} $ is the approximate vector difference between the camera and the plane of the drogue center (where we install LEDs) along the $ z_{\rm{c}} $ axis under the camera coordinate frame as shown in Fig. 3.
\begin{figure}[hbt!]
	\centering
	\includegraphics[width=.6\textwidth]{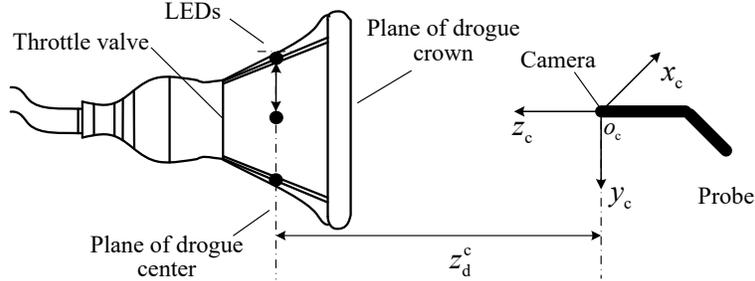}
	\caption{Schematic diagram of docking depth.}
\end{figure} 
In order to simplify the calculation, by eliminating some unimportant variables, Eq. (4) is decomposed into lateral channel and longitudinal channel.
\begin{itemize}[itemindent=10pt]
	\item [$\cdot  $ ] 
	In the $ x_{\rm{c}}-z_{\rm{c}} $ plane, the three degrees of freedom are $ {v}_{{\rm{d}},x}^{\rm{c}} $, $ {v}_{{\rm{d}},z}^{\rm{c}} $, $ {\omega}_{{\rm{d}},y}^{\rm{c}} $ with $ {v}_{{\rm{d}},y}^{\rm{c}}=0 $, $ {\omega}_{{\rm{d}},x}^{\rm{c}}=0 $. We can obtain 
	\begin{equation}
		\dot{e}_{x}=-\frac{{v}_{{\rm{d}},x}^{\rm{c}} }{{z}_{\rm{d}}^{\rm{c}} } +\frac{e_{x}{v}^{\rm{c}}_{{\rm{d}},z} }{{z}_{\rm{d}}^{\rm{c}} } -(1+e_{x}^{2} ){\omega}_{{\rm{d}},y}^{\rm{c}}.
	\end{equation}
	\item [$\cdot  $ ] 
	In the $ y_{\rm{c}}-z_{\rm{c}} $ plane, the three degrees of freedom are $ {v}_{{\rm{d}},y}^{\rm{c}} $, $ {v}_{{\rm{d}},z}^{\rm{c}} $, $ {\omega}_{{\rm{d}},x}^{\rm{c}} $ with $ {v}_{{\rm{d}},x}^{\rm{c}}=0 $, $ {\omega}_{{\rm{d}},y}^{\rm{c}}=0 $. We can obtain 
	\begin{equation}
		\dot{e}_{y}=-\frac{{v}_{{\rm{d}},y}^{\rm{c}} }{{z}_{\rm{d}}^{\rm{c}} } +\frac{e_{y}{v}_{{\rm{d}},z}^{\rm{c}} }{ {z}_{\rm{d}}^{\rm{c}} } -(1+e_{y}^{2} ){\omega}_{{\rm{d}},x} ^{\rm{c}}.
	\end{equation}
\end{itemize}
\subsection{Visual servo model based on the decoupling model}

The receiver can be taken as a rigid body, the camera's linear velocity is equal to the vector sum of the velocity of the receiver's center of the mass and the velocity of the camera rotating around the center of the mass under the tanker coordinate frame, i.e.,
\begin{equation}
	\begin{split}
		\mathbf{v}_{\rm{c}}^{\rm{t}}
		&=\boldsymbol{{\omega}}_{\rm{r}}^{\rm{t}}\times \mathbf{p}_{\rm{c}}^{\rm{r}}+\mathbf{{v}}_{\rm{r}}^{\rm{t}}
	\end{split}
\end{equation}
where $ \mathbf{{v}}_{\rm{c}}^{\rm{t}}=[{v}^{\rm{t}}_{{\rm{c}},x} {v}^{\rm{t}}_{{\rm{c}},y} {v}^{\rm{t}}_{{\rm{c}},z}]^{\rm{T}} $ is the velocity of the camera, $ \mathbf{{v}}_{\rm{r}}^{\rm{t}}=[{v}^{\rm{t}}_{{\rm{r}},x} {v}^{\rm{t}}_{{\rm{r}},y} {v}^{\rm{t}}_{{\rm{r}},z}]^{\rm{T}} $ is the linear velocity of the receiver, $ \boldsymbol{{\omega}}_{\rm{r}}^{\rm{t}}=[{\omega}^{\rm{t}}_{{\rm{r}},x} {\omega}^{\rm{t}}_{{\rm{r}},y} {\omega}^{\rm{t}}_{{\rm{r}},z}]^{\rm{T}} $ is the angular velocity of the receiver and $ \mathbf{p}_{\rm{c}}^{\rm{r}}=$ [ $ x_{\rm{c}}^{\rm{r}} $ $ y_{\rm{c}}^{\rm{r}} $ $ z_{\rm{c}}^{\rm{r}} $ ] $^{\rm{T}} $ is the camera's coordinate under receiver coordinate frame as shown in Fig. 1.

According to the definitions of frames, the transfer matrix from the receiver coordinate frame to the tanker coordinate frame is 
\begin{equation}
	\mathbf{R} \rm{_{t/r}}=\textbf{I}_{3}.
\end{equation}
With Eq. (8), Eq. (7) becomes
\begin{equation}
	\begin{split}
		\mathbf{{v}}_{\rm{c}}^{\rm{t}}
		&=\boldsymbol{{\omega}}_{\rm{r}}^{\rm{t}}\times \mathbf{p}_{\rm{c}}^{\rm{r}}+\mathbf{{v}}_{\rm{r}}^{\rm{t}}\\
		&=\mathbf{R} \rm{_{t/r}}\boldsymbol{{\omega}}_{\rm{r}}^{\rm{r}}\times \mathbf{p}_{\rm{c}}^{\rm{r}}+\mathbf{R} \rm{_{t/r}}\mathbf{{v}}_{\rm{r}}^{\rm{r}}\\
		&=\boldsymbol{{\omega}}_{\rm{r}}^{\rm{r}}\times \mathbf{p}_{\rm{c}}^{\rm{r}}+\mathbf{{v}}_{\rm{r}}^{\rm{r}}
	\end{split}
\end{equation}
where $ \boldsymbol{{\omega}}_{\rm{r}}^{\rm{r}}= $ [ $ \tilde{p} $ $ \tilde{q} $ $ \tilde{r} $ ] $ ^{\rm{T}} $ is the components of velocity $ \boldsymbol{{\omega}}_{\rm{r}}^{\rm{t}} $ on each axis of the receiver coordinate system , and $ \mathbf{{v}}_{\rm{r}}^{\rm{r}} $ = [ $ \tilde{u} $ $ \tilde{v} $ $ \tilde{w} $ ] $ ^{\rm{T}} $ is the components of velocity $ \mathbf{{v}}_{\rm{r}}^{\rm{t}} $ on each axis of the receiver coordinate system. 
Based on Eq. (14), the camera's linear velocity $ \textbf{v}_{\rm{c}}^{\rm{re}} $ can be expressed as

\begin{equation}
	\begin{split}
	\mathbf{{v}}_{\rm{c}}^{\rm{re}}
	&=\mathbf{R}_{\rm{re/t}}\mathbf{{v}}_{\rm{c}}^{\rm{t}} \\
	&=\mathbf{R} \rm{_{re/t}}(\boldsymbol{{\omega}}_{\rm{r}}^{\rm{r}}\times \mathbf{p}_{\rm{c}}^{\rm{r}}+\mathbf{{v}}_{\rm{r}}^{\rm{r}}).
  \end{split}
\end{equation}
Then, according to Ref. \cite{beard2012small} ,
\begin{equation}
	\left\{\begin{aligned}
		{\tilde{u}} &= {\tilde{V}^{\rm{g}}_{\rm{r}}}\cos \tilde{\alpha} \cos \tilde{\beta}  \\ 
		{\tilde{v}} &= {\tilde{V}^{\rm{g}}_{\rm{r}}}\sin \tilde{\beta}  \\ 
		{\tilde{w}} &= {\tilde{V}^{\rm{g}}_{\rm{r}}}\sin \tilde{\alpha} \cos \tilde{\beta}  \\ 
	\end{aligned}\right.  
\end{equation}
Eq. (10) becomes 
\begin{equation}
	\left\{ \begin{aligned}
		{v}_{{\rm{c}},x}^{\rm{re}} &={\tilde{V}^{\rm{g}}_{\rm{r}}}\sin \tilde{\beta}+x_{\rm{c}}^{\rm{r}}\tilde{r}-z_{\rm{c}}^{\rm{r}}\tilde{p} \hfill \\
	{v}_{{\rm{c}},y}^{\rm{re}} &= {\tilde{V}^{\rm{g}}_{\rm{r}}}\sin \tilde{\alpha} \cos \tilde{\beta}+ y_{\rm{c}}^{\rm{r}}\tilde{p}-x_{\rm{c}}^{\rm{r}}\tilde{q} \hfill\\
	{v}_{{\rm{c}},z}^{\rm{re}} &= {\tilde{V}^{\rm{g}}_{\rm{r}}}\cos \tilde{\alpha} \cos \tilde{\beta}+ z_{\rm{c}}^{\rm{r}}\tilde{q}- y_{\rm{c}}^{\rm{r}}\tilde{r} .\hfill 
	\end{aligned}  \right.
\end{equation}
Based on the decoupling conditions for longitudinal channel  $\tilde{p} = \tilde{r} = 0 $, $\tilde{\beta}  = 0$ and omitting the higher-order item, Eq. (12) becomes
\begin{equation}
	\left\{ \begin{aligned}
	 v_{{\rm{c}},x}^{\text{re}} &=  x_{\rm{c}}^{\rm{r}}\tilde r - z_{\rm{c}}^{\rm{r}}\tilde p \\
		 v_{{\rm{c}},y}^{\text{re}} &=  - x_{\rm{c}}^{\rm{r}}\tilde q   \hfill \\
 v_{{\rm{c}},z}^{\text{re}} &= {{\tilde V}_{\rm{r}}^{\rm{g}}}+z_{\rm{c}}^{\rm{r}}\tilde q.   \hfill 
	\end{aligned}  \right. 
\end{equation}

Combining Eq. (1) and Eq. (13), the longitudinal model is obtained
\begin{equation}
	\left\{ \begin{aligned}
	 	\boldsymbol{\dot{\tilde{\rm{x}} }} _{\rm{rlon}} &=\rm{\mathbf{A} }_{rlon} \tilde{\boldsymbol{\rm{x}} } _{\rm{rlon}}+\rm{\mathbf{B} }_{rlon}\tilde{\boldsymbol{\rm{u}} } _{\rm{rlon}}\hfill \\
	\mathbf{ v}_{{\text{rlon}}}^{{\text{re}}} &\triangleq\left[ {\begin{array}{*{20}{c}}
			{ v_{{\rm{c}},y}^{\rm{re}}} \\ 
			{ v_{{\rm{c}},z}^{\rm{re}}} 
	\end{array}} \right] = \underbrace {\left[ {\begin{array}{*{20}{c}}
		0&0&0&0&0&{ - {x_{\rm{c}}^{\rm{r}}}} \\ 
		0&0&0&1&0&{{z_{\rm{c}}^{\rm{r}}}} 
\end{array}} \right]}_{\mathbf{C}_{{\text{rlon}}}}\tilde{\boldsymbol{\rm{x}} } _{\rm{rlon}}
	\end{aligned}  \right.
\end{equation}
and the lateral model is denoted by
\begin{equation}
	\left\{ \begin{aligned}
			\dot{\tilde{\boldsymbol{\rm{x}} } } _{\rm{rlat}} &=\rm{\mathbf{A} }_{rlat} \tilde{\boldsymbol{\rm{x}} } _{\rm{rlat}}+\rm{\mathbf{B} }_{rlat}\tilde{\boldsymbol{\rm{u}} } _{\rm{rlat}}  \hfill \\
			 v_{{\text{rlat}}}^{{\text{re}}} &\triangleq  v_{{\rm{c}},x}^{\rm{re}} = \underbrace{\left[ {\begin{array}{*{20}{l}}
				0&0&0&0&{ -}z_{\rm{c}}^{\rm{r}}&x_{\rm{c}}^{\rm{r}}
		\end{array}} \right]}_{\textbf{C}_{{\text{rlat}}}}\tilde{\boldsymbol{\rm{x}} } _{\rm{rlat}}
	\end{aligned}  \right.
\end{equation}

\subsection{Visual servo controller design}
Based on the models (5), (6), (14) and (15), the controller design is divided into the following two sub-problems.

\textbf{Problem 1.} (Lateral channel controller design). For system (15), design $ \tilde{\boldsymbol{\rm{u}} } _{\rm{rlat}} $, such that the lateral image tracking error converges to zero, i.e.,  $ e_{x}(t)\to 0  $ as $ t\to\infty  $.

\textbf{Problem 2.} (Longitudinal channel controller design). For system (14), design $ \tilde{\boldsymbol{\rm{u}} } _{\rm{rlon}} $, such that the the longitudinal image tracking error converges to zero and the probe approaches the drogue, i.e., $ e_{y}(t)\to 0  $, $ z_{\rm{d}}^{\rm{c}}(t)\to 0 $ as $ t\to\infty  $.	
\begin{itemize}[itemindent=10pt]
	\item [1)]\textit{ Outer loop controller design}:

	\begin{itemize}[itemindent=10pt]
	
	\item [$\cdot  $ ] 
	Lateral channel controller design. For \textit{Problem 1}, consider that the change of $ \omega_{{\rm{d}},y}^{\rm{c}} $ is small in the docking process which can be ignored. Then the desired velocity for $ {v}_{{\rm{d}},x}^{\rm{c}} $ is designed as		\end{itemize}

	\begin{equation}
		v_{{\rm{d}},x\rm{des}}^{\text{c}}=k_{1}e_{x}.
	\end{equation}
	With it, if $ 	v_{{\rm{d}},x}^{\text{c}}=	v_{{\rm{d}},x\rm{des}}^{\text{c}} $, then Eq. (10) becomes
	\begin{equation}
		\dot{e}_{x}=-\lambda_{1}e_{x}
	\end{equation}
	where $ \lambda_{1}=\frac{k_{1}- v_{{\rm{d}},z}^{\text{c}}}{z_{\rm{d}}^{\rm{c}}} $ and $ k_{1} $  is chosen as $ k_{1}>{\rm{max}}(v_{{\rm{d}},z}^{\rm{c}}) $. In this case, we have $ \mathop {\lim}\limits_{t \to \infty }\left| e_{x}(t) \right|=0$.

	\begin{itemize}[itemindent=10pt]
	\item [$\cdot  $ ] 
	Longitudinal channel controller design. 
	For \textit{Problem 2}, consider that the change of $ \omega_{{\rm{d}},x}^{\rm{c}} $ is small in the docking process which can be ignored. Then the desired velocity for $ {v}_{{\rm{d}},y}^{\rm{c}} $ is designed as \end{itemize}
	\begin{equation}
		{v}_{{\rm{d}},y\rm{des}}^{\rm{c}}=k_{2}e_{y}.
	\end{equation}
	With it, if $ 	v_{{\rm{d}},y}^{\text{c}}=	v_{{\rm{d}},y\rm{des}}^{\text{c}} $, Eq. (6) becomes 
	\begin{equation}
		\dot{e}_{y}=-\lambda_{2}e_{y}
	\end{equation}
where $ \lambda_{2}=\frac{k_{2}-{v}_{{\rm{d}},z}^{\rm{c}}}{z_{\rm{d}}^{\rm{c}}} $ and $ k_{2} $  is chosen as $ k_{2}>\text{max}({v}_{{\rm{d}},z}^{\rm{c}}) $. In this case, we have $ \mathop {\lim}\limits_{t \to \infty }\left| e_{y}(t) \right|=0$. Besides, the desired velocity for $ v_{{\rm{d}},z}^{\text{c}} $ is designed as
\begin{equation}
	v_{{\rm{d}},z\rm{des}}^{\text{c}}= -k_{3}z_{\rm{d}}^{\rm{c}}  
\end{equation}

With it, if $v_{{\rm{d}},z}^{\text{c}}=	v_{{\rm{d}},z\rm{des}}^{\text{c}} $ and $ k_{3}>0 $, we have $ \mathop {\lim}\limits_{t \to \infty }\left| z_{\rm{d}}^{\rm{c}}(t) \right|=0 $. In the following, two improvements are made on (25). First, in order to prevent the receiver too fast to pass the drogue, the term ``$ -k_{4}\left | e_{x}  \right | -k_{5} \left | e_{y}  \right |  $'' is introduced into (25) to adjust the speed as shown in the following
\begin{equation}
	v_{{\rm{d}},z\rm{des}}^{\text{c}} = {-k_{3}z_{\rm{d}}^{\rm{c}}  -k_{4}\left | e_{x}  \right | -k_{5} \left | e_{y}  \right | }.
\end{equation}
If $ \left | e_{x}  \right | $, $ \left | e_{y}  \right | $ is large, the term ``$ -k_{4}\left | e_{x}  \right | -k_{5} \left | e_{y}  \right |  $''  can slow down the speed to avoid overshooting. Secondly, an offset $ a>0 $ is added as a constant velocity to guarantee the receiver still fly at a certain forward velocity after $ z_{\rm{d}}^{\rm{c}}(t)=0 $, because the receiver should hit to open the throttle valve with a certain relative speed at the time of docking as shown in Fig. 3, but the camera cannot capture the LEDs after  $ z_{\rm{d}}^{\rm{c}} (t)=0 $.  The final controller is designed as
\begin{equation}
	v_{{\rm{d}},z\rm{des}}^{\text{c}} ={\rm{min}}\left ( {-k_{3}z_{\rm{d}}^{\rm{c}}  -k_{4}\left | e_{x}  \right | -k_{5} \left | e_{y}  \right | }, a\right )
\end{equation}
where $ k_{3}, k_{4}, k_{5}>0 $ and $ a<0 $.
 	\end{itemize}
 
	\begin{itemize}[itemindent=10pt]
	
	\item [$\cdot  $ ] 
	Synthesis. Up to now, the outer loop controllers for longitudinal and lateral channel are finished and the desired velocity for $ \textbf{v}^{\rm{c}}_{\rm{d}} $ is obtained, denoted as   \end{itemize}
\begin{equation}
	\textbf{v}_{{\rm{d}},\rm{des}}^{\rm{c}}=\left[\begin{matrix}
		
		v_{{\rm{d}},x\rm{des}}^{\text{c}} &
		v_{{\rm{d}},y\rm{des}}^{\text{c}} &
		v_{{\rm{d}},z\rm{des}}^{\text{c}}
	\end{matrix}\right]^{\rm{T}}. 
\end{equation}
In practice, since $ \textbf{v}_{\rm{d}}^{\rm{c}}=\textbf{v}_{\rm{c}}^{\rm{re}}-\textbf{v}_{\rm{d}}^{\rm{re}} $
, we can only control $ \textbf{v}_{\rm{c}}^{\rm{re}} $ rather than $ \textbf{v}_{\rm{d}}^{\rm{re}} $, because $ \textbf{v}_{\rm{d}}^{\rm{re}} $ is the dynamics of the drogue. So, roughly, we will use $ \textbf{v}_{\rm{c},des}^{\rm{re}}=\textbf{v}_{{\rm{d}},\rm{des}}^{\rm{c}}  $ by  taking $ \textbf{v}_{\rm{d}}^{\rm{re}} $ as a disturbance, which is expressed in a decoupled form as
\begin{equation}
	{\textbf{v}}^{\rm{re}}_{\rm{c,des}}=\begin{bmatrix}
		{\textbf{v}}_{\rm{rlon},des}^{\rm{re}}  \\
		{{v}}_{\rm{rlat},des}^{\rm{re}}
	\end{bmatrix} 
\end{equation}
where ${\textbf{v}}^{\rm{re}}_{\rm{rlon},des}=$ [${v}^{\rm{c}}_{{\rm{d}},y\rm{des}}$ 
	$ {v}^{\rm{c}}_{{\rm{d}},z\rm{des}} $ ]$ ^{\rm{T}} $,  $ {v}^{\rm{re}}_{\rm{rlat},des}={v}^{\rm{c}}_{{\rm{d}},x\rm{des}} $. The task of the inner loop controller is to make $ {\textbf{v}}_{\rm{c}}^{\rm{re}} $ track $ {\textbf{v}}_{\rm{c,des}}^{\rm{re}} $ and reject complex disturbances.

	\begin{itemize}[itemindent=10pt]
			\item [2)] \textit{Inner loop controller design}: 
	\item [$\cdot  $ ] 
	Longitudinal channel controller design. 
	Define the tracking error as $ \textbf{e}_{\rm{rlon}}={\textbf{v}}_{\rm{rlon}}^{\rm{re}}-{\textbf{v}}^{\rm{re}}_{\rm{rlon},des} $, an integrator is used here to reject complex disturbances, which is defined as 	\end{itemize}
	\begin{equation}
		\mathbf{q}_{\rm{rlon}}=\int_{0}^{t} \mathbf{e}_{\rm{rlon}}(\tau)\rm{d\tau}=\int_0^{\emph{t}}({\mathbf{v}}^{\rm{re}}_{\rm{rlon}}(\mathbf{\tau})-{\mathbf{v}}^{\rm{re}}_{\rm{rlon},des}(\mathbf{\tau}))d\tau.
	\end{equation}
	Putting Eq. (25) to Eq. (1) yields
	\begin{equation}
		\begin{bmatrix}
			\dot{\tilde{\mathbf{x} }} _{\rm{rlon}} \\
			\dot{\mathbf{q} } _{\rm{rlon}}
		\end{bmatrix}=\begin{bmatrix}
			\mathbf{A}_{\rm{rlon}} & \mathbf{0}_{6\times 2} \\
			\mathbf{C}_{\rm{rlon}}&\mathbf{0}_{2\times 2}
		\end{bmatrix}\begin{bmatrix}
			\tilde{\mathbf{x}}_{\rm{rlon}}\\
			\mathbf{q}_{\rm{rlon}}
		\end{bmatrix}+\begin{bmatrix}
			\mathbf{B}_{\rm{rlon}}\\
			\mathbf{0}_{2\times2}
		\end{bmatrix}\mathbf{\tilde{u}}_{\rm{rlon}}-\begin{bmatrix}
			\mathbf{0}_{6\times 2} \\
			\mathbf{I}_{2\times 2}
		\end{bmatrix}\mathbf{{v}}_{{\rm{rlon},des}}^{\rm{re}}.
	\end{equation}
	The controller is designed as follows
	\begin{equation}
		\tilde{\mathbf{u}}_{\rm{rlon}}=-\mathbf{K}_{x1}\tilde{\mathbf{x}}_{\rm{rlon}}-\mathbf{K}_{e1}\mathbf{q}_{\rm{rlon}}
	\end{equation}
	where $ \mathbf{K}_{x1}\in {\mathbf{\mathbb{R}}^{2 \times 6}}, \mathbf{K}_{e1}\in {\mathbf{\mathbb{R}}^{2 \times 2}} $. Then a cost function is designed to obtain $\mathbf{K}_{x1}  $ and $\mathbf{K}_{e1}$, which is defined as follows
	\begin{equation}
		J(\tilde{\mathbf{u}}_{\rm{rlon}})=\underset{\mathbf{K}_{x1},\mathbf{K}_{e1} }{ \rm{argmin}}\int_{0}^{\infty } \left \{ \begin{bmatrix}
			\mathbf{\tilde{x}}_{\rm{rlon}} ^{\rm{T}} & \mathbf{q}_{\rm{rlon}}^{\rm{T}}
		\end{bmatrix}\mathbf{Q}_{\rm{rlon}}\begin{bmatrix}
			\mathbf{\tilde{x}}_{\rm{rlon}}\\
			\mathbf{q}_{\rm{rlon}}
		\end{bmatrix} +\mathbf{\tilde{u}}^{\rm{T}}_{\rm{rlon}}\mathbf{R}_{\rm{rlon}} \mathbf{\tilde{u}}_{\rm{rlon}} \right \}{\rm{d}}t
	\end{equation}
where $ \mathbf{Q}_{\rm{rlon}} \succeq 0 $ and $ \mathbf{R}_{\rm{rlon}}\succ 0 $ are weighting matrices.
\begin{itemize}[itemindent=10pt]
	\item [$\cdot  $ ] 
	 Lateral channel controller design. Similarly, define the tracking error as $ {e}_{\rm{rlat}}={v}_{\rm{rlat}}^{\rm{re}}-{v}_{{\rm{rlat},des}}^{\rm{re}} $ and an integrator is defined as 	\end{itemize} 
	\begin{equation}
		{{q}}_{\rm{rlat}}=\int_{0}^{t} e_{{\text{rlat}}}(\tau){\rm{d\tau}}=\int_{0}^{t}({v}^{{\rm{re}}}_{\rm{rlat}}(\mathbf{\tau})-{v}^{\rm{re}}_{\rm{rlat},des}(\mathbf{\tau}))\rm{d}\tau.
	\end{equation}
Putting Eq. (29) to Eq. (1) yields
\begin{equation}
	\begin{bmatrix}
		\dot{\tilde{\mathbf{x} }} _{\rm{rlat}} \\
		\dot{{q} } _{\rm{rlat}}
	\end{bmatrix}=\begin{bmatrix}
		\mathbf{A}_{\rm{rlat}} & \mathbf{0}_{6\times 2} \\
		\mathbf{C}_{\rm{rlat}}&\mathbf{0}_{2\times 2}
	\end{bmatrix}\begin{bmatrix}
		\tilde{\mathbf{x}}_{\rm{rlat}}\\
		{q}_{\rm{rlat}}
	\end{bmatrix}+\begin{bmatrix}
		\mathbf{B}_{\rm{rlat}}\\
		\mathbf{0}_{2\times2}
	\end{bmatrix}\mathbf{\tilde{u}}_{\rm{rlat}}-\begin{bmatrix}
		\mathbf{0}_{6\times 2} \\
		\mathbf{I}_{2\times 2}
	\end{bmatrix}{v}_{{\rm{rlat},des}}^{\rm{re}}.
\end{equation}
	The controller is designed as 
	\begin{equation}
		\tilde{\mathbf{u}}_{\rm{rlat}}=-\mathbf{K}_{x2}\tilde{\mathbf{x}}_{\rm{rlat}}-\mathbf{K}_{e2}{q}_{\rm{rlat}}
	\end{equation}
	where $ \mathbf{K}_{x2}\in {\mathbb{R}}^{{2 \times 6}}, \mathbf{K}_{e2}\in {\mathbb{R}}^{{2}} $. And the cost function is defined as
	\begin{equation}
		J(\tilde{\mathbf{u}}_{\rm{rlat}})=\underset{\mathbf{K}_{x2},\mathbf{K}_{e2} }{ \rm{argmin}}\int_{0}^{\infty } \left \{ \begin{bmatrix}
			\mathbf{\tilde{x}}_{\rm{rlat}} ^{\rm{T}} & {q}_{\rm{rlat}}^{\rm{T}}
		\end{bmatrix}\mathbf{Q}_{\rm{rlat}}\begin{bmatrix}
			\mathbf{\tilde{x}}_{\rm{rlat}}\\
			{q}_{\rm{rlat}}
		\end{bmatrix} +\mathbf{\tilde{u}}_{\rm{rlat}}^{\rm{T}}\mathbf{R}_{\rm{rlat}} \mathbf{\tilde{u}}_{\rm{rlat}} \right \}{\rm{d}}t
	\end{equation}
where $ \mathbf{Q}_{\rm{rlat}} \succeq 0 $ and $ \mathbf{R}_{\rm{rlat}}\succ 0 $ are weighting matrices.

	\begin{itemize}[itemindent=10pt]
	
	\item [$\cdot  $ ] 
	Synthesis. Up to now, based on the models (1), (14) and (15), and under the \textit{Assumptions 1-3}, the final controller is obtained as follows, 	\end{itemize}
	\begin{equation}
		\left\{\begin{aligned} 
			{v}_{{\rm{d}},x\rm{des}}^{\rm{c}}&=k_{1}e_{x} \\  
			{v}_{{\rm{d}},y\rm{des}}^{\rm{c}}&=k_{2}e_{y} \\
			{v}_{{\rm{d}},z\rm{des}}^{\rm{c}}&={\rm{min}}\left ( {-k_{3}z_{\rm{d}}^{\rm{c}}   -k_{4}\left | e_{x}  \right | -k_{5} \left | e_{y}  \right | }, a\right) \\
			\mathbf{\tilde{u}}_{\rm{rlon}}&=-\mathbf{K}_{x1}\tilde{\mathbf{x} }_{\rm{rlon}}-\mathbf{K}_{e1}\mathbf{q}_{\rm{rlon}}\\
			\mathbf{\tilde{u}}_{\rm{rlat}}&=-\mathbf{K}_{x2}\tilde{\mathbf{x} }_{\rm{rlat}}-\mathbf{K}_{e2}{q}_{\rm{rlat}}  
		\end{aligned}\right. 
	\end{equation}
	where $ k_{i}>0 $, $ i=1,2,...,5 $ are control gains. The structure of the IBVS controller is shown in Fig. 4.

\begin{figure}[hbt!]
	\centering
	\includegraphics[width=1\textwidth]{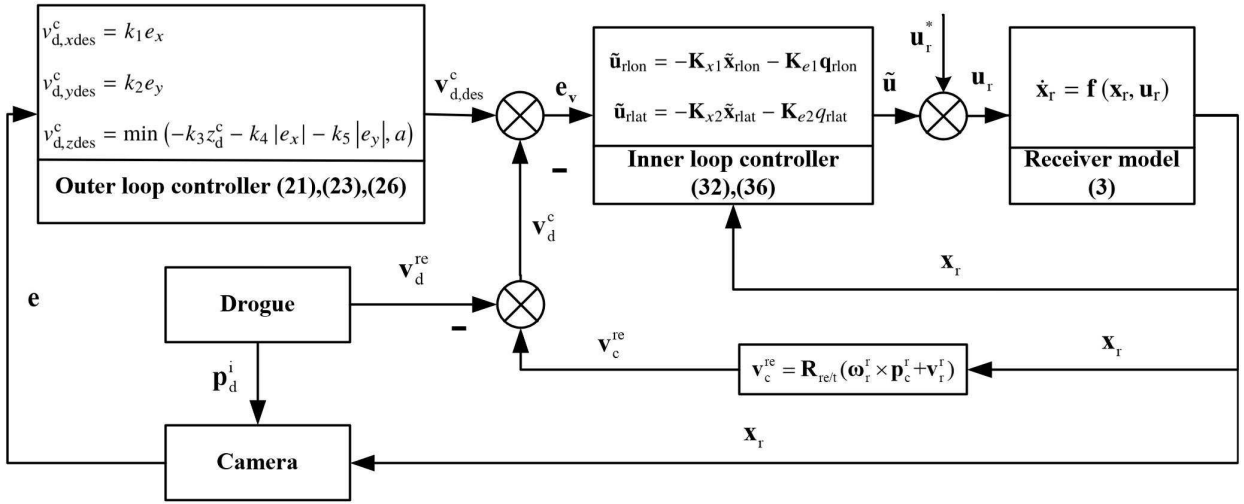}
	\caption{The structure of the IBVS controller.}
\end{figure}

\section{Simulation and Verification}
In this section, different disturbances such as aerodynamic disturbance, bow wave effect and pose estimation error are considered in the simulation to verify whether the controller designed is valid. An introduction to the paper has been uploaded to the url:  \url{https://www.youtube.com/watch?v=g-pkieLL-RI} or \url{http://rfly.buaa.edu.cn/index.html#/home}.
\subsection{Simulation environment}
A MATLAB/SIMULINK based simulation environment with a 3D virtual-reality display is adopted to simulate AAR docking, whose detailed information about the modeling procedure, model parameters, and simulation environment can be referred to Refs. \cite{7738351,DAI2016448}.

\subsection{Simulation results}
Different initial conditions and disturbances are set in the simulation environment to demonstrate the proposed IBVS method's effectiveness. Besides, the pose estimation error error is considered to make a comparison between IBVS and PBVS.

\begin{itemize}[itemindent=10pt]
	\item [1)]\textit{Aerodynamic Disturbance}   
	
	The parameters of the outer loop controller are set as seen in the Table 1 below. At first, different intensity of the atmospheric disturbance are added.The intensity of the atmospheric disturbance is negatively correlated with the probability of the turbulence intensity being exceeded. In general, the intensity of the atmospheric disturbance of level \uppercase\expandafter{\romannumeral1} means that the probability of the turbulence intensity being exceeded is ${10^{ - 1}}$ and the maximum velocity of atmospheric disturbance is about $5$ feet per second. The level \uppercase\expandafter{\romannumeral2} means that the probability of the turbulence intensity being exceeded is ${10^{ - 2}}$ and the maximum velocity of atmospheric disturbance is about $7$ feet per second\cite{moorhouse1980us}.
	\begin{table}[h]
		\centering
		\begin{tabular}{|l|c|c|c|c|c|c|}\hline
			Parameters&$ k_{1} $&$ k_{2} $&$ k_{3} $&$ k_{4} $&$ k_{5} $\\\hline
			Values&1&2&0.3&3&1\\\hline
		\end{tabular}
		\caption{Parameters of the outer loop}
		\label{tab:Parameters of the outer loop}
	\end{table} 
	
In Fig. 4(a), the intensity of aerodynamic disturbance is set at level \uppercase\expandafter{\romannumeral1} while the intensity of aerodynamic disturbance is set at level \uppercase\expandafter{\romannumeral2} in Fig. 4(b), we can find that although the receiver's movement trajectory fluctuates with the increase of the wind disturbance intensity, the success of docking is ensured while the trajectory is relatively smooth due to the effect of the `PI' controller. Then, a bow wave effect is added in Fig. 5 in which the parameters of the outer loop controller are adjusted as shown in Table 2. Noteworthy, although the image tracking error fluctuates at the beginning, it can still guarantee the success of docking.	
\begin{table}[h]
	\centering
	\begin{tabular}{|l|c|c|c|c|c|c|}\hline
		Parameters&$ k_{1} $&$ k_{2} $&$ k_{3} $&$ k_{4} $&$ k_{5} $\\\hline
		Values&3&3&0.5&5&2\\\hline
	\end{tabular}
	\caption{Parameters of the outer loop under the bow wave effect}
	\label{tab:Parameters of the outer loop under the bow wave effect}
\end{table}
	\begin{figure}[htbp]
		\centerline{\includegraphics[width=7in]{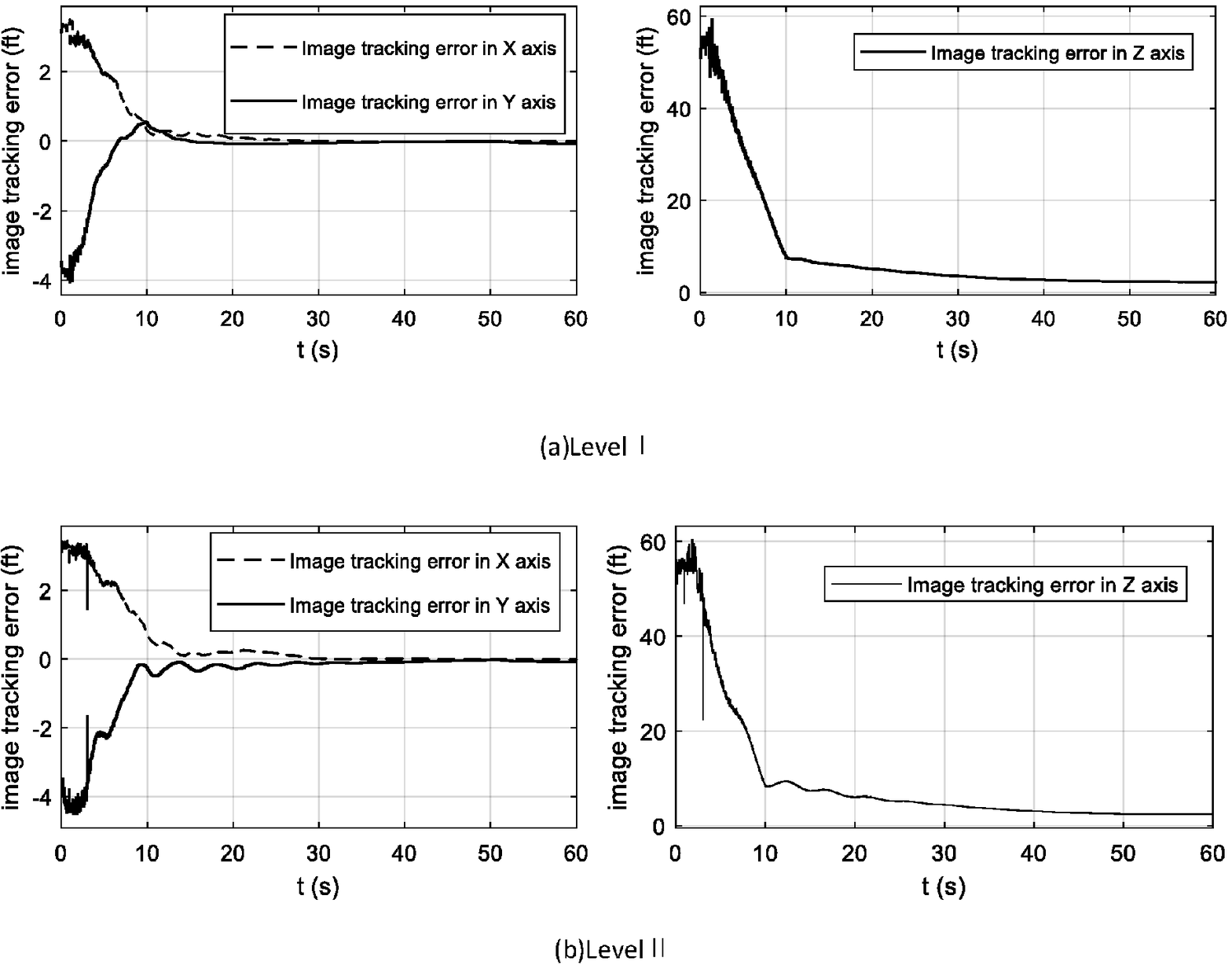}}
		\caption{Image tracking error under different aerodynamic disturbances}
		\label{fig6}
	\end{figure}

	\begin{figure}[htbp]
		\centerline{\includegraphics[width=7in]{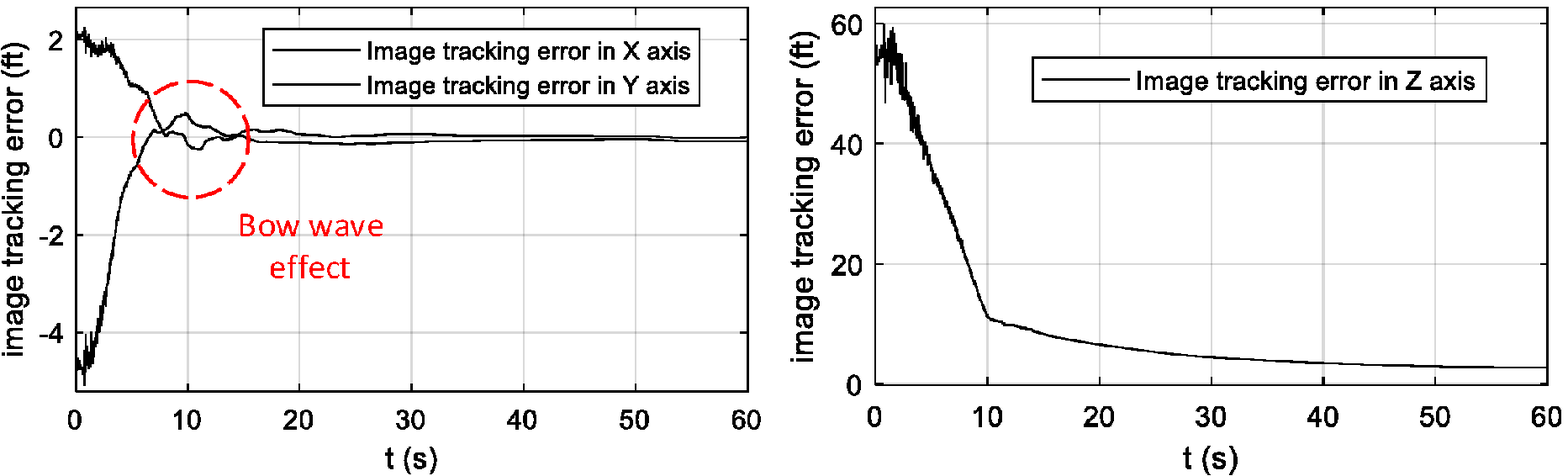}}
		\caption{Image tracking error under bow wave effect}
		\label{fig7}
	\end{figure}
	\begin{figure}[htbp]
		\centerline{\includegraphics[width=7in]{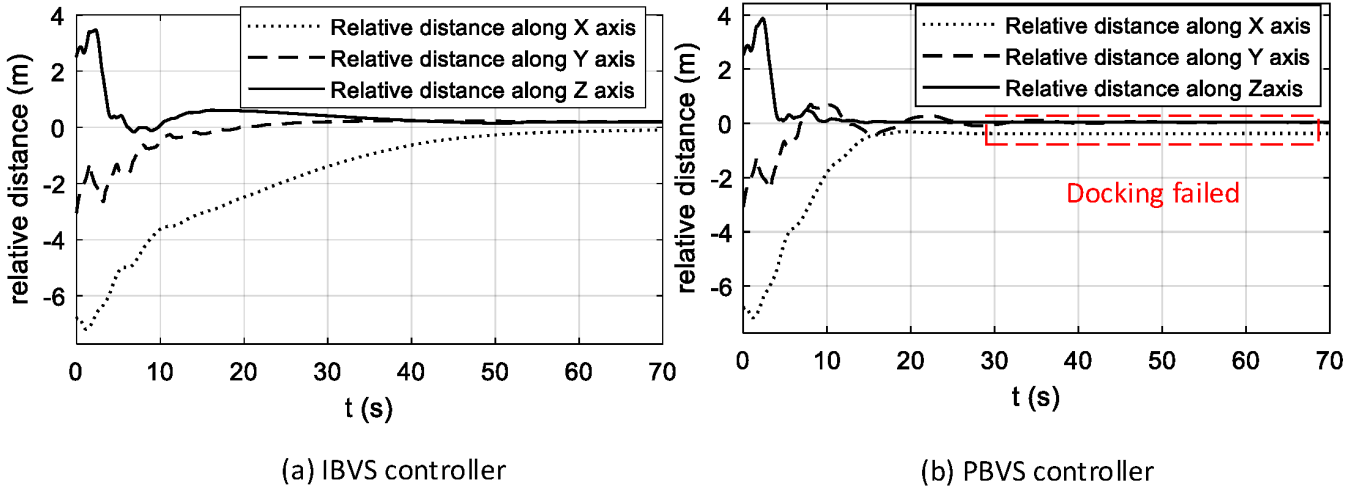}}
		\caption{Image tracking error under position measurement error}
		\label{fig8}
	\end{figure}
	\item [2)]\textit{Position Measurement Error}
	
	Consider that the pose estimation error after the camera installation affects the docking control in two aspects: the camera installation position does not change, but the distance measurement may exist error, or the measurement is precise,
	but the actual position of the camera is deviated from the original measurement position. To simulate the situation, $ \Delta \mathbf{p} _{\rm{c}}^{\rm{r}} =$ [  1  0 -0.5 ] $ ^{\rm{T}} $  is added to both IBVS and PBVS controller as the position error as shown in Fig. 6. It can be observed that the IBVS controller is able to perform a successful docking while the PBVS cannot. 
	\item [3)]\textit{Discussion}

	The PBVS takes the inaccurate 3D relative position as the outer loop feedback measurement. A PI controller can make the measurement error zero. However, real error exists so the docking fails finally. On the other hand, the IBVS takes the accurate relative 2D image error as the outer loop feedback measurement. The other error exists in the inner loop which is taken as constant disturbances, or the depth error does not determine the success in docking too much like the 2D image error owing to the design in (22). As a result, the IBVS `PI' controller is more robust against measure error than the PBVS `PI' controller.

\end{itemize}
\section{Conclusions}
This paper proposes an IBVS model for probe-drogue refueling and a docking control method. Concretely, the IBVS control for the outer loop and the LQR control for the inner loop are designed to improve the robustness of the docking. Simulations show that the proposed control method is robust to achieve a successful docking control under aerodynamic disturbances, including the bow wave effect, and pose estimation error. Here, the throttle, elevator, aileron and rudder are control inputs of the system. However, the low-level controller always has been existed and cannot be accessed for safety in practice. Therefore, a further research on an existing low-level controller where the velocity is adopted as the control input is deserved to study. Often, the velocity controller aims at receiver's position control. But the velocity controller designed here is required to aim at the probe tip position control related to the receiver's attitude. This is full of challenge.

\bibliography{sample}

\end{document}